\documentclass[10pt,twocolumn,letterpaper]{article}

\newcommand{\fig}[1]{Fig. \ref{#1}}

\newcommand{\sect}[1]{Sect. \ref{#1}}

\usepackage{multirow}
\usepackage{mdwtab}

\usepackage{wacv}
\usepackage{times}
\usepackage{epsfig}
\usepackage{graphicx}
\usepackage{amsmath}
\usepackage{amssymb}

\wacvfinalcopy % *** Uncomment this line for the final submission

\ifwacvfinal\fi
\begin{document}

%%%%%%%%% TITLE
\title{Spatiotemporal Stacked Sequential Learning for Pedestrian Detection}

% Authors at the same institution
%\author{First Author \hspace{2cm} Second Author \\
% Authors at different institutions
\author{Alejandro Gonz{\'a}lez$^{1,2}$ \hspace{0.5cm} Sebastian Ramos$^{1,2}$ \hspace{0.5cm} David V\'azquez$^1$\\ \hspace{1cm} Antonio M. L\'opez$^{1,2}$ \hspace{1cm} Jaume Amores$^{1,3}$\\ \\ 
$^1$ Computer Vision Center, Barcelona \hspace{2cm} $^2$ Universitat Aut\`{o}noma de Barcelona\\ $^3$ United Technologies Research Center\\ \\ {\tt\small \{agalzate,sramosp,dvazquez,antonio,jaume\}@cvc.uab.es}}

%\author{Alejandro Gonz{\'a}lez \\
%%Computer Vision Center, Barcelona\\
%Universitat Aut\`{o}noma de Barcelona\\
%{\tt\small agalzate@cvc.uab.es}
%\and
%Sebastian Ramos \\
%Computer Vision Center\\
%{\tt\small sramosp@cvc.uab.es}
%\and
%David V{\'a}zquez \\
%Computer Vision Center\\
%{\tt\small dvazquez@cvc.uab.es}
%\and
%Antonio M. L{\'o}pez \\
%%Computer Vision Center\\
%Autonomous University of Barcelona\\
%{\tt\small antonio@cvc.uab.es}
%\and
%Jaume Amores \\
%Computer Vision Center\\
%{\tt\small jaume@cvc.uab.es}
%}

\maketitle
\ifwacvfinal\thispagestyle{empty}\fi

%%%%%%%%% ABSTRACT
\begin{abstract}
   Pedestrian classifiers decide which image windows contain a pedestrian. In practice, such classifiers provide a relatively high response at neighbor windows overlapping a pedestrian, while the responses around potential false positives are expected to be lower. An analogous reasoning applies for image sequences. If there is a pedestrian located within a frame, the same pedestrian is expected to appear close to the same location in neighbor frames. Therefore, such a location has chances of receiving high classification scores during several frames, while false positives are expected to be more spurious. In this paper we propose to exploit such correlations for improving the accuracy of base pedestrian classifiers. In particular, we propose to use two-stage classifiers which not only rely on the image descriptors required by the base classifiers but also on the response of such base classifiers in a given spatiotemporal neighborhood. More specifically, we train pedestrian classifiers using a stacked sequential learning (SSL) paradigm. We use a new pedestrian dataset we have acquired from a car to evaluate our proposal at different frame rates. We also test on a well known dataset: Caltech. The obtained results show that our SSL proposal boosts detection accuracy significantly with a minimal impact on the computational cost. Interestingly, SSL improves more the accuracy at the most dangerous situations, {\ie} when a pedestrian is close to the camera.
\end{abstract}

%%%%%%%%% BODY TEXT
\section{Introduction}

Localizing humans in images is key for applications such as video surveillance, avoiding pedestrian-to-vehicle collisions, collecting statistics of players or athletes in sport videos, etc. Developing a reliable vision-based pedestrian detector is a very challenging task with more than a decade of history by now. As a result, a plethora of features, models, and learning algorithms, have been proposed to develop the pedestrian classifiers which are at the core of pedestrian detectors \cite{Geronimo:2013}. 

The research for boosting the accuracy of pedestrian classifiers has followed different lines. Some authors have researched image descriptors well-suited for pedestrians ({\eg}, HOG \cite{Dalal:2005}, HOG+LBP \cite{Wang:2009}, HOG+CSS+HOF \cite{Walk:2010}, OppHOG \cite{Rao:2011}, Haar+EOH \cite{Geronimo:2010b}, Integral Channels \cite{Dollar:2009b}, Macrofeatures \cite{Nam:2011}), others have researched different image modalities  ({\eg},  appearance + motion \cite{Wojek:2009}, appearance+depth+motion \cite{Enzweiler:2011}), others have focused on the pedestrian model ({\eg}, deformable multi-component part-based models \cite{Felzenszwalb:2010, Ramanan:2011, Cho:2012}, multi-resolution \cite{Park:2010, Benenson:2012}), others on the classification architecture ({\eg}, HOG-SVM/LRF-MLP cascades \cite{Oliveira:2010}, Haar + EOH-AdaBoost cascades with meta-stages \cite{Chen:2008}, random forest of HOG+LBP-SVMs \cite{Marin:2013b}), and others in the process of collecting good samples for training ({\eg}, generative approach \cite{Enzweiler:2008}, active learning \cite{Abramson:2005}, virtual-world data with domain adaptation \cite{Vazquez:2013b}). 

The outcome of each of the above mentioned proposals is a pedestrian classifier, termed here as {\em base classifier}, which determines if a given image window contains a pedestrian or background. In practice, such classifiers provide a relatively high response at neighbor windows overlapping a pedestrian, while the responses around potential false positives are expected to be lower. Note that, in fact, non-maximum suppression (NMS) is usually performed as last detection stage in order to reduce multiple detections arising from the same pedestrian to a single one. An analogous reasoning applies for image sequences. If there is a pedestrian located within a frame, the same pedestrian is expected to appear close to the same location in neighbor frames. Therefore, such a location has chances of receiving high classification scores during several frames, while false positives are expected to be more spurious. In fact, this may allow removing such undesired spurious by the use of a tracker. 

In this paper we propose to exploit such expected {\em response correlations} for improving the accuracy of the classification stage itself. In other words, instead of only exploiting spatiotemporal coherence by means of general post-classification stages like NMS and tracking, we propose to add such a type of reasoning in the classification stage itself as well. In particular, we propose to use a two-stage classification strategy which not only rely on the image descriptors required by the base classifiers, but also on the response of the own base classifiers in a given spatiotemporal neighborhood. More specifically, we train pedestrian classifiers using a stacked sequential learning (SSL) paradigm \cite{Cohen:2005}.

Temporal SSL involves the analysis of window volumes. The different types of temporal volumes can be potentially useful for different applications depending on the motion of the camera and the targets of interest, as well as the working frame rate and the targets size. In this paper, we are specially interested in on-board pedestrian detection within urban scenarios. Therefore, camera and targets are in movement. Accordingly, in this paper we test our SSL approach for a fixed neighborhood ({\ie}, fixed spatial window coordinates across frames) and for an scheme relying on an ego-motion compensation approximation ({\ie}, varying spatial window coordinates across frames).  Moreover, in order to assess the dependency of the results with respect to the frame rate, we acquired our own pedestrian dataset at 30fps by normal driving in an urban scenario. This new dataset is used as main guide for our experiments, but we also complement our study with other challenging dataset publicly available: Caltech.

In this paper we start by using a competitive baseline in pedestrian detection \cite{Dollar:2012}, namely a holistic base classifier based on HOG+LBP features and linear SVM. Note that HOG/linear-SVM is the core of more sophisticated pedestrian detectors as the popular deformable part-based model (DPM) \cite{Felzenszwalb:2010}. Moreover, HOG with LBP are also used as base descriptors of multi-modal multi-view pedestrian models \cite{Enzweiler:2011}, and HOG+LBP/linear-SVM has been used for classifiers with occlusion handling \cite{Wang:2009, Marin:2013}, as well as for acting as node experts in random forest ensembles \cite{Marin:2013b}. In addition, it has recently been shown that HOG+LBP/linear-SVM approaches are well suited for domain adaptation \cite{Vazquez:2013b}. Altogether, we think that HOG+LBP/linear-SVM is a proper baseline to start assessing our proposal. Moreover we have extended this baseline with the HOF \cite{Walk:2010} motion descriptor that complements the appearance and texture features of the baseline.

Overall, the obtained results show that our spatiotemporal SSL proposal boosts detection accuracy significantly. Especially, when the pedestrians are close to the camera, {\ie} in the most critical situations. Therefore, encouraging to augment the study for other pedestrian base classifiers as well as other object categories. 

The rest of the paper is organized as follows. In \sect{sec:rw} we review some works related to our proposal. Section \ref{sec:ssl} briefly introduces the SSL paradigm. In \sect{sec:sslpd} we develop our proposal. Section \ref{sec:er} presents the experiments carried out to assess our spatiotemporal SSL, and discuss the obtained results. Finally, \sect{sec:conclusion} draws our main conclusions.

%------------------------------------------------------------------------
\section{Related work}
\label{sec:rw}

The use of motion patterns as image descriptors was already proposed as an extension of spatial Haar-like filters for video surveillance applications (static zenital camera) \cite{Viola:2003, Cui:2007, Jones:2008} and for detecting human visual events \cite{Ke:2005}. In these cases, original spatial Haar-like filters were extended with a temporal dimension. Popular HOG descriptor was also extended to encode temporal information for detecting humans \cite{Dalal:2006b}, in this case using optical flow to compensate motion. In the same spirit the histograms of flow (HOF) were also introduced for detecting pedestrians \cite{Walk:2010}. In all cases motion information was complemented with appearance information ({\ie}, Haar/HOG for luminance and/or color channels).

In contrast with these approaches, our proposal does not involve to compute new temporal image descriptors as new features for the classification process. As we will see, we use the responses of a given base classifier in neighbor frames as new features for our SSL classifier. In fact, our proposal can also be applied to base classifiers that already incorporate motion features. Therefore, the reviewed literature and our proposal are complementary strategies.

Focusing on single frames, it has been recently shown how pedestrian detection accuracy can be boosted by analyzing the image area surrounding potential pedestrian detections. In particular, \cite{Ding:2012, Chen:2013} follow an iterative process that uses contextual features of several orders ({\eg}, involving co-occurences) for progressively enhancing the response of base classifiers for true pedestrians and lowering it for hallucinatory ones. Our SSL proposal does not require new image descriptors of pedestrian surroundings and is not iterative, which makes it inherently faster. Moreover, we treat equally spatial and temporal response correlations, {\ie}, under the SSL paradigm, giving rise to a more straightforward method. 

Finally, we would like to clarify that our SSL proposal is not a substitute for NMS and tracking post-classification stages. What we expect is to allow these stages to produce more accurate results by increasing the accuracy of the classification stage. For instance, tracking must be used for predicting pedestrian intentions \cite{Schneider:2013}, thus, if less false positives reach the tracker, we can reasonably expect to obtain more reliable pedestrian trajectories and so guessing intentions in the very short time this information is required ({\ie}, around a quarter of second before a potential collision).

\section{Stacked sequential learning (SSL)}
\label{sec:ssl}

\begin{figure*}[t!]
\centering
\includegraphics[width=0.9\textwidth]{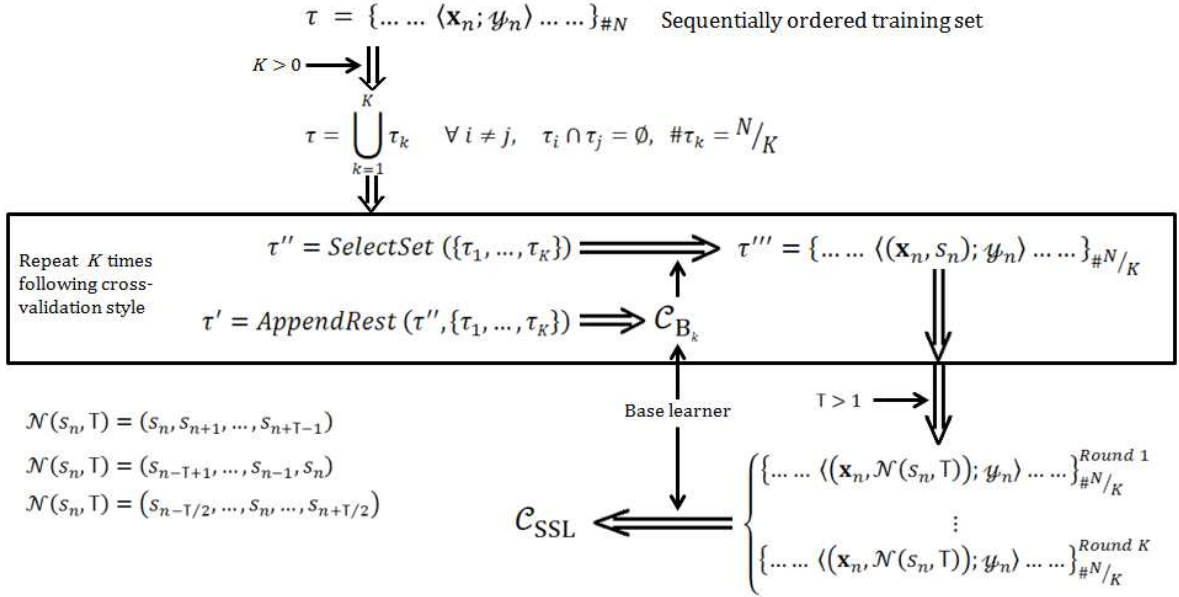}
\caption{SSL learning. See main text in \sect{sec:ssl} for details.}
\label{fig:ssl}
\end{figure*}

Stacked sequential learning (SSL) was introduced by Cohen {\etal} \cite{Cohen:2005} with the aim of improving base classifiers when the data to be processed has some sort of sequential order. In particular, given a data sample to be classified, the core intuition is to consider not only the features describing the sample but also the response of the base classifier in its neighbor samples. Figure \ref{fig:ssl} summarizes the SSL learning process that we explain in more detail in the rest of this section. 

Let $\tau$ be an ordered training sequence of cardinality $N$. The SSL approach involves to select a sub-sequence for training a base classifier, ${\cal C}_B$, and the rest to apply ${\cal C}_B$ and so training the SSL classifier, ${\cal C}_{SSL}$. If this is done once, then the final classifier ${\cal C}_{SSL}$ would be trained with less than $N$ samples. Thus, to avoid this, it is followed a cross-validation style were $\tau$ is divided in $K>0$ disjoint sub-sequences, $\tau = \cup_{k=1}^K \tau_k \wedge i \neq j \Rightarrow \tau_i\cap\tau_j=\emptyset$, and $K$ rounds are performed by using a different subset each round to test the ${\cal C}_{B_k}$ and the rest of subsets for training this ${\cal C}_{B_k}$. At the end of the process, joining the $K$ sub-sequences processed by the corresponding ${\cal C}_{B_k}$, we can have $N$ {\em augmented} training samples for learning ${\cal C}_{SSL}$. $k=1$ means to train the $C_B$ and $C_SSL$ on the same training set, without actually doing partitions.

Let us explain what means {\em augmented} training samples. The elements of  $\tau$, {\ie}, the initial training samples, are of the form $<{\bf x_n};y_n>$, where ${\bf x_n}$ is a vector of features with associated label $y_n$. Therefore, the elements of each sub-sequence $\tau_k$ are of the same form. As we have mentioned before, during each round $k$ of the cross-validation-style process, a sub-sequence $\tau''$ is selected among $\{\tau_1,\ldots,\tau_K\}$, while the rest are appended together to form a sub-sequence $\tau'$. From $\tau'$ it is learned ${\cal C}_{B_k}$ and applied to $\tau''$ to obtain a new $\tau'''$. The elements of $\tau'''$ are of the form $<({\bf x_n},s_n);y_n>$, where we have augmented the feature ${\bf x_n}$ with the classifier score $s_n={\cal C}_{B_k}({\bf x_n})$. Therefore, after the $K$ rounds, we have a training set of $N$ samples of the form $<({\bf x_n},s_n);y_n>$. It is at this point when we can introduce the concept of neighbor scores into the learning process. In particular, the final training samples are of the form $<({\bf x_n},{\cal N}(s_n,T));y_n>$, where ${\cal N}(s_n,T)$ denotes a neighborhood of size $T>1$ anchored to the sample $n$. For instance, ${\cal N}(s_n,T)=(s_{n-T+1},\ldots,s_{n-1},s_n)$ is a {\em past} neighborhood, ${\cal N}(s_n,T)=(s_n,s_{n+1},\ldots,s_{n+T-1})$ is a {\em future} neighborhood, and ${\cal N}(s_n,T)=(s_{n-\frac{T}{2}},\ldots,s_n,\ldots,s_{n+\frac{T}{2}})$ is a {\em centered} neighborhood, which are analogous concepts to the ones of filtering, extrapolation and smoothing, resp., used in the classical tracking nomenclature.

%------------------------------------------------------------------------
\section{SSL for pedestrian detection}
\label{sec:sslpd}

\begin{figure*}[t!]
\centering
\includegraphics[width=0.9\textwidth]{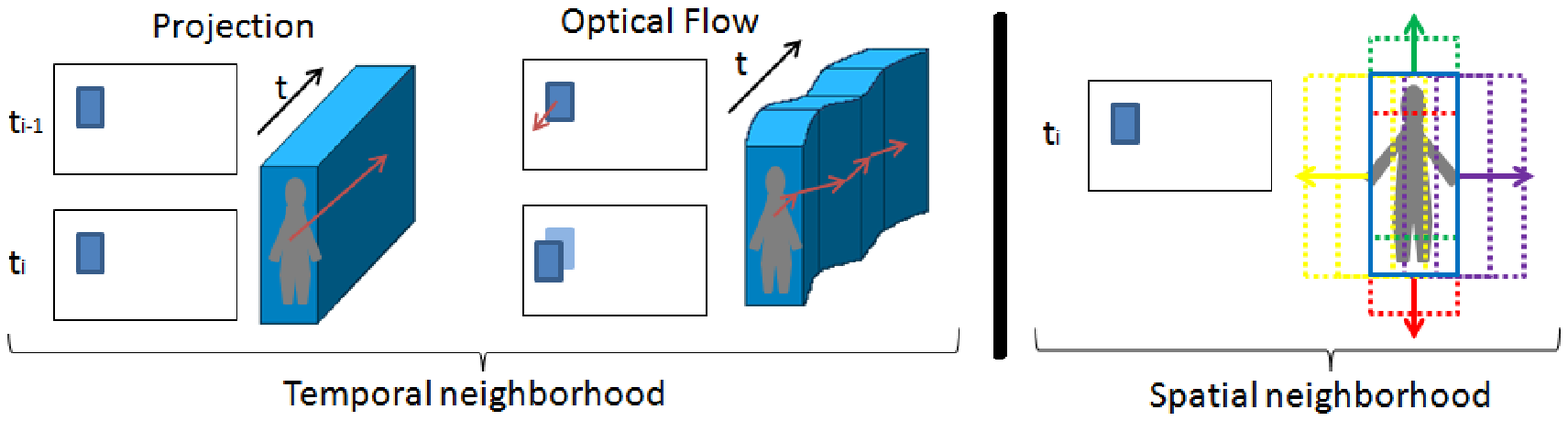}
\caption{Different types of neighborhood for SSL. See main text in \sect{ssec:spatiotemporal} for details.}
\label{fig:neighborhoods}
\end{figure*}

In this section, without loosing generality, we will assume the use of the {\em past neighborhood} (\sect{sec:ssl}) to illustrate and explain our SSL approach. From the viewpoint of the processing of image sequences, this means to use previous images to do detection in the current one ({\ie}, in the last one acquired when processing directly from a camera). Actually there is no need to save the previous images. The detection scores of the neighbouring windows, that were already computed, are enough to compute the current SSL descriptor making the computation of SSL very computational efficient.

\subsection{Spatiotemporal neighborhoods for SSL}
\label{ssec:spatiotemporal}

For object detection in general and for pedestrian detection in particular, applying SSL starts by defining which are the neighbors of a given window under analysis. In learning time, such a window will correspond either to the bounding box of a labeled pedestrian or to a rectangular chunk of the background. In operation time ({\ie}, testing), such a window will correspond to a candidate generated by a pyramidal sliding window scheme or any other candidate selection method. In this paper we assume the processing of image sequences and, consequently, we propose the use of a spatiotemporal neighborhood.

Temporal SSL involves the analysis of window volumes. Therefore, there are several possibilities to consider (see \fig{fig:neighborhoods}). Let us term as $W_f$ the set of coordinates defining an image window in frame $f$, and ${\bf V}_f=\mbox{vol}(\cup_{t=0}^{T-1}W_{f-t})$ the window volume defined by a temporal neighbor of $T$ frames. The simplest volume is obtained by assuming fixed locations across frames, which we term as {\em projection} approach. In other words,  $W_f=W_{f-1}=...=W_{f-(T-1)}$. Another possibility consists in building volumes taking into account motion information. For instance, $W_f=W_{f-1}+t_{OF(W_{f-1})}$, where $t_{OF(W_{f-1})}$ is a 2D translation defined by considering the {\em optical flow} contained in $W_{f-1}$, and $'+'$ stands for summation to all coordinates defining $W_{f-1}$.  

Spatial SSL involves the analysis of windows spatially overlapping the window of interest (see \fig{fig:neighborhoods}). For instance, we can fix a 2D displacement $\Delta=(\delta_x,\delta_y)$ and $n_x$ displacements in the $x$ axis, to the left and to the right, an analogously for the $y$ axis given a $n_y$ number of up and down displacements.

Our proposal combines both ideas, {\ie}, the temporal volumes and the spatial overlapping windows, in order to define the spatiotemporal neighborhood required by SSL (\sect{sec:ssl}).

\subsection{SSL training}
\label{ssec:SSLtraining}

As usual, we assume an image sequence with labeled pedestrians ({\ie}, using bounding boxes) for training. Negative samples for training are obtained by random sampling of the same images, of course, these samples cannot highly overlap labeled pedestrians. The cross-validation-style rounds of SSL (\sect{sec:ssl}) are performer with respect to the images of the sequence, not with respected to the set of labeled pedestrians and negative samples as it may suggest the straightforward application of SSL (note that pedestrian/negative labels are for individual windows not for full images). Moreover, as we have seen in \sect{ssec:spatiotemporal}, the neighborhood relationship is not only temporal but spatial too. The training process is divided in two stages. First, we train the auxiliary classifiers (${\cal C}_{B_k}$) as usual using three bootstraping rounds. Then we train the SSL classifier (using final ${\cal C}_{B_k}$ as auxiliary), again we run three bootstrapping rounds for obtaining the final classifier (${\cal C}_{SSL}$).

Using the full training dataset, we also assume the training of a base classifier ${\cal C}_{B}$. Another possibility is to understand the different ${\cal C}_{B_k}$ as the result of a bagging procedure and ensemble them to obtain ${\cal C}_{B}$. Without loosing generality, in this paper we have focused on the former approach. 

\begin{figure*}%[t!]
\centering
\includegraphics[width=\textwidth]{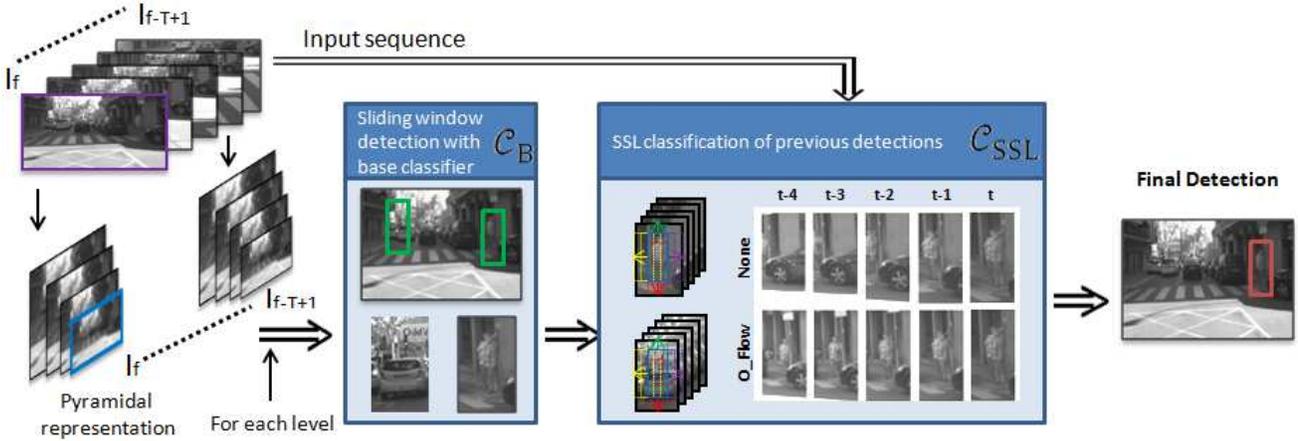}
\caption{Two-stage pedestrian detection based on SSL. See main text in \sect{ssec:SSLdetection} for details.}
\label{fig:SSLdetection}
\end{figure*}

\subsection{SSL detector}
\label{ssec:SSLdetection}

The proposed pedestrian detection pipeline is shown in \fig{fig:SSLdetection}. As we can see there are two main stages. The first stage basically consists in a classical pedestrian detection method relying on the learned base classifier ${\cal C}_{B}$. In \fig{fig:SSLdetection} we have illustrated the idea for a pyramidal sliding window approach, but using other candidate selection approaches is also possible. Detections at this stage are just considered as potential ones. Then, the second stage applies the spatiotemporal SSL classifier, ${\cal C}_{SSL}$, to such potential detections in order to reject or keep them as final detections. 

There are some details worth to mention. First, the usual non-maximum suppression (NMS) step included in pedestrian detectors is not performed for the output of the first stage, but it is done for the output of the second stage. Second, for ensuring that true pedestrians reach the second stage, we apply a threshold on ${\cal C}_{B}$ such that it guarantees a very high detection rate even having a very high rate of false positives. In our experiments this usually implies that while the ${\cal C}_{B}$ processes hundred of thousands windows (for pyramidal sliding window), ${\cal C}_{SSL}$ only process a few thousands. Third, although in \fig{fig:SSLdetection} we show pyramids of images for a temporal neighborhood of T frames, what we actually keep from frame to frame are the already computed features, so that we compute them only once. However, this depends on the type of temporal neighborhood we use (\sect{ssec:spatiotemporal}). For instance, using projection style no feature are needed to keep ({\ie}, keeping the classification scores is enough). However, if we use optical flow we may need to compute features in previous frames if the window under consideration does not map to a location where they were already computed.

%------------------------------------------------------------------------
\section{Experimental results}
\label{sec:er}

\paragraph {\bf Protocol.} As evaluation methodology we follow the de-facto Caltech standard for pedestrian detection  \cite{Dollar:2012}, {\ie} we plot curves of false positives per image (FPPI) {\vs} miss rate. The miss rate average in the range of $10^{-2}$ to $10^{0}$ FPPI is taken as indicative of each detector accuracy, {\ie.} the lower the better. 
Moreover, during testing we consider three different subset based on the pedestrian height. {\em Near} subset include pedestrians with height equal or higher than 75 pixels, {\em medium} subset include pedestrian between 50 and 75 pixel height. Finally we group the two previous subset in the {\em reasonable} subset (height $>=$ 50 pixels).

\begin{table*}[ht]
  \centering
\caption{Evaluation of SSL over different datasets, frame rates and pedestrian sizes. For FPPI $\in [0.01,1]$, the miss rate average $\%$ is indicated.}
\vspace{0.5cm}
  % Create the table  
{\scriptsize 
\begin{tabular}{|r|c|l|c|c|c|}\hline 
 {\bf Dataset} & {\bf FPS} & {\bf Experiment}  & {\bf Near} & {\bf Medium} & {\bf Reasonable}\\ \hline \hline

\multirow{10}{*}{OursDS} & Any & Base: HOG+LBP & 39.71 & 50.83 & 45.91 \\\hlx{c{2-6}} 
 & \multirow{3}{*}{3} & SSL(Base) Proj. - OptFl. & 36.03 -  36.72 & 50.01 - 50.04 & 44.40 - 44.02 \\\hlx{c{3-6}}
 &   & Base+HOF & 47.98 & 56.65 & 50.88 \\\hlx{c{3-6}}  
 &   & SSL(Base+HOF) Proj. & 37.62 & 52.21 & 45.47 \\\hlx{c{2-6}}
 & \multirow{3}{*}{10} & SSL(Base) Proj. - OptFl. & 35.49 - 34.79 & 50.22 - 49.42 & 43.56 - 42.10 \\\hlx{c{3-6}} 
 &  & Base+HOF & 39.24 & 52.37 & 42.43 \\\hlx{c{3-6}}  
 &  & SSL(Base+HOF) Proj. & 29.42 & 44.62 & 37.13 \\\hlx{c{2-6}}
 & \multirow{3}{*}{30} & SSL(Base) Proj. - OptFl. & 34.18 - 34.01 & 49.84 - 48.04 & 42.90 - 41.73 \\\hlx{c{3-6}}  
 &  & Base+HOF & 37.81 & 53.39 & 38.78 \\\hlx{c{3-6}}   
 &  & SSL(Base+HOF) Proj. & 27.37 & 46.53 & 35.85 \\ \hline  \hline
\multirow{4}{*}{Caltech} & \multirow{4}{*}{25} & Base & 45.4 & 82.3 & 59.4 \\\hlx{c{3-6}}
 &  & SSL(Base) Proj. - OptFl. & 40.6 - 38.9 & 81.2 - 80.4 & 59.4 - 57.6 \\\hlx{c{3-6}} 
 &  & Base+HOF & 33.8 & 78.4 & 52.9 \\\hlx{c{3-6}}  
 &  & SSL(Base+HOF) Proj. & 32.0 & 77.1 & 51.6 \\ \hline 
\end{tabular}
}
\label{SSL:table:Results}
\end{table*}

\paragraph {\bf Our own dataset (OurDS).} Since the temporal axis is important for the SSL classifier, we acquired our own dataset to be sure we have stable 30 fps sequences. The sequences were acquired on-board under normal urban driving conditions. The images are monochrome and of $480\times960$ pixels. We used a 4mm focal length lens, so providing a wide field of view. We drove during 30 minutes approximately, giving rise to a sequence of around 60,000 frames. Then, using steps of 10 frames we annotated all the pedestrians. This turns out in 7,900 annotated pedestrians, 5,400 reasonable and non occluded. We have divided the video sequence into three sequential parts, the first one for training, the last one for testing, in the middle we have leaved a gap for avoiding testing and training with the same persons. Overall we train with 3,600 reasonable pedestrians, and test on 1,300 reasonable ones.

\paragraph {\bf Caltech dataset.} We have also used other popular dataset acquired on-board. The Caltech dataset \cite{Dollar:2012}, which contain 3,700 reasonable pedestrians for training. 

\paragraph {\bf Base detectors.} For the experiments presented in this section we use our own implementation of HOG and LBP features, which provides significant better results than the one proposed in \cite{Wang:2009}, {\ie}, removing the occlusion handling reasoning. Moreover, using TV-L1 \cite{Zach:2007} for computing optical flow, we obtain HOF features \cite{Walk:2010} as well. These features complement HOG and LBP by motion information. We call Base to the HOG+LBP/Linear-SVM and Base+HOF to the HOG+LBP+HOF/Linear-SVM.

\begin{figure*}[t!]
\centering
{
\includegraphics[width=0.3\linewidth]{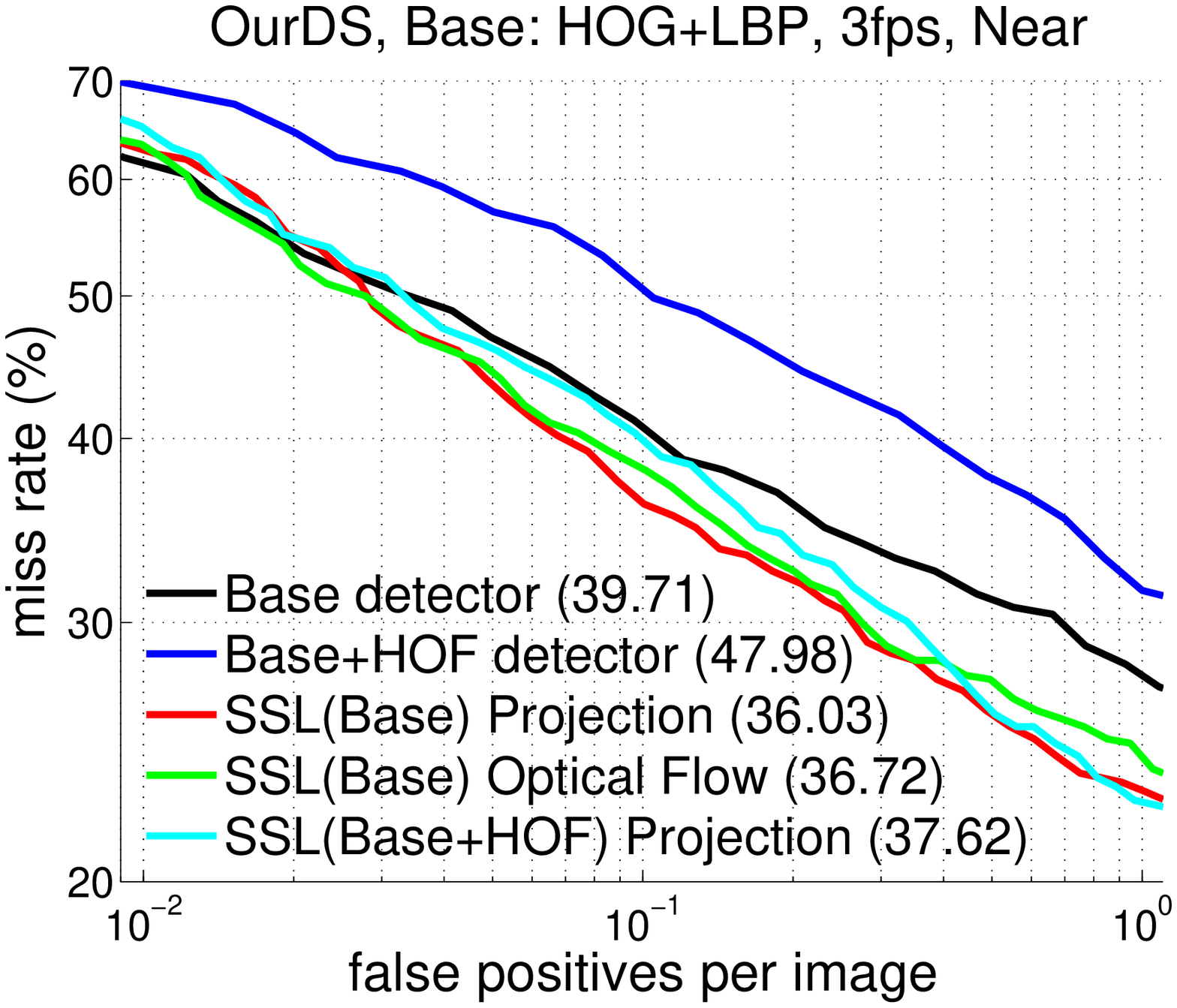}
\includegraphics[width=0.3\linewidth]{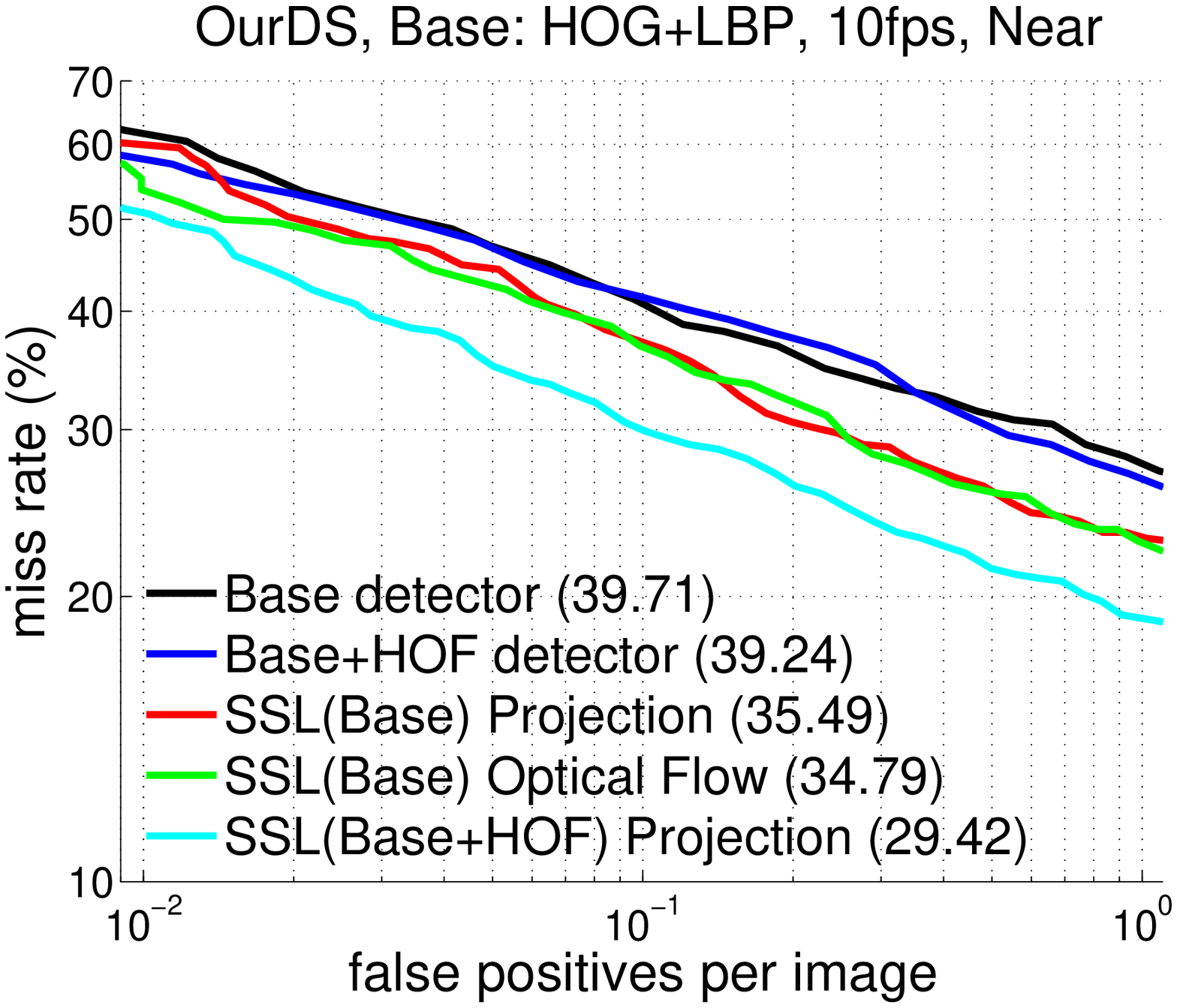}
\includegraphics[width=0.3\linewidth]{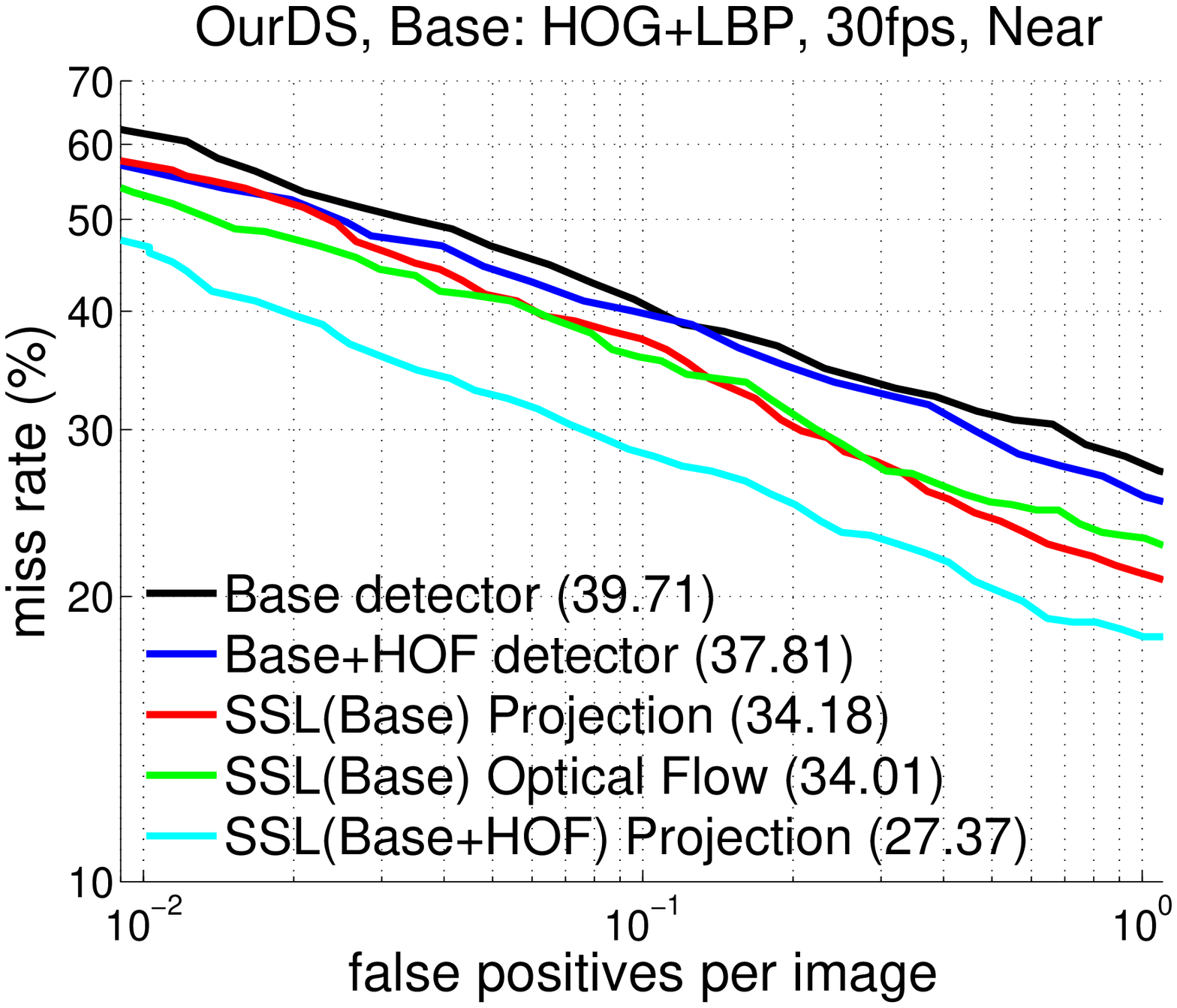}\\
\includegraphics[width=0.3\linewidth]{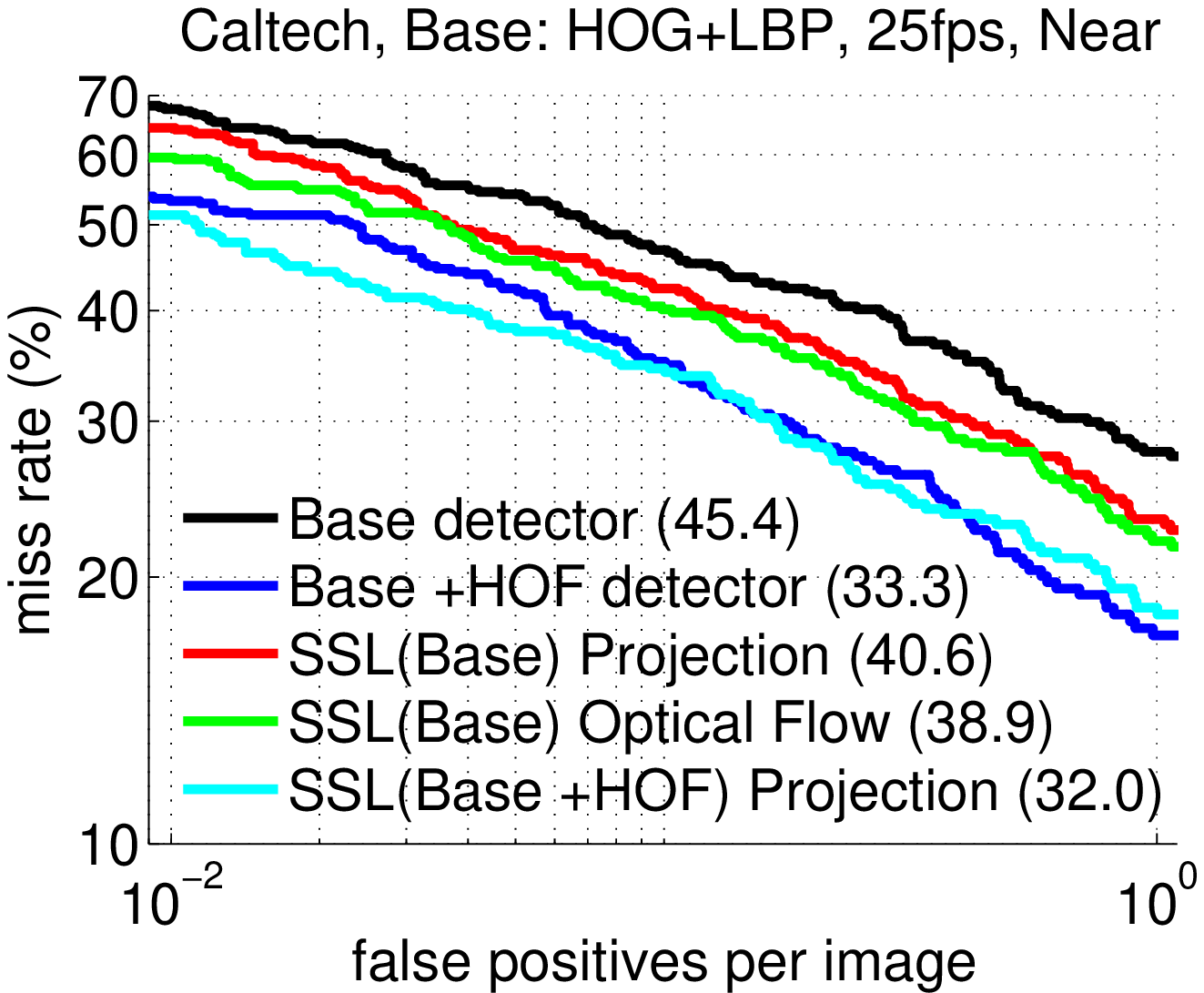}
\includegraphics[width=0.3\linewidth]{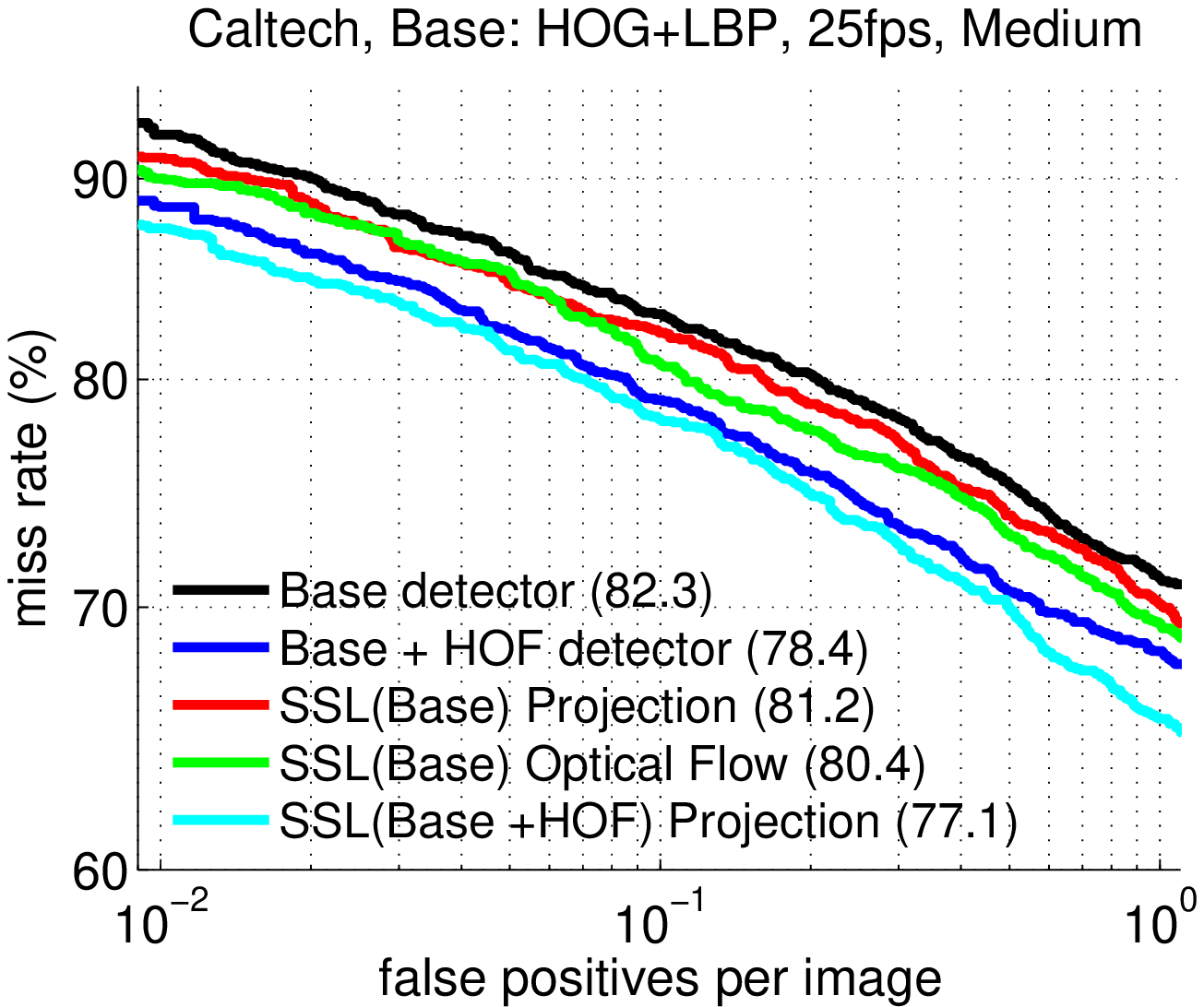}
\includegraphics[width=0.3\linewidth]{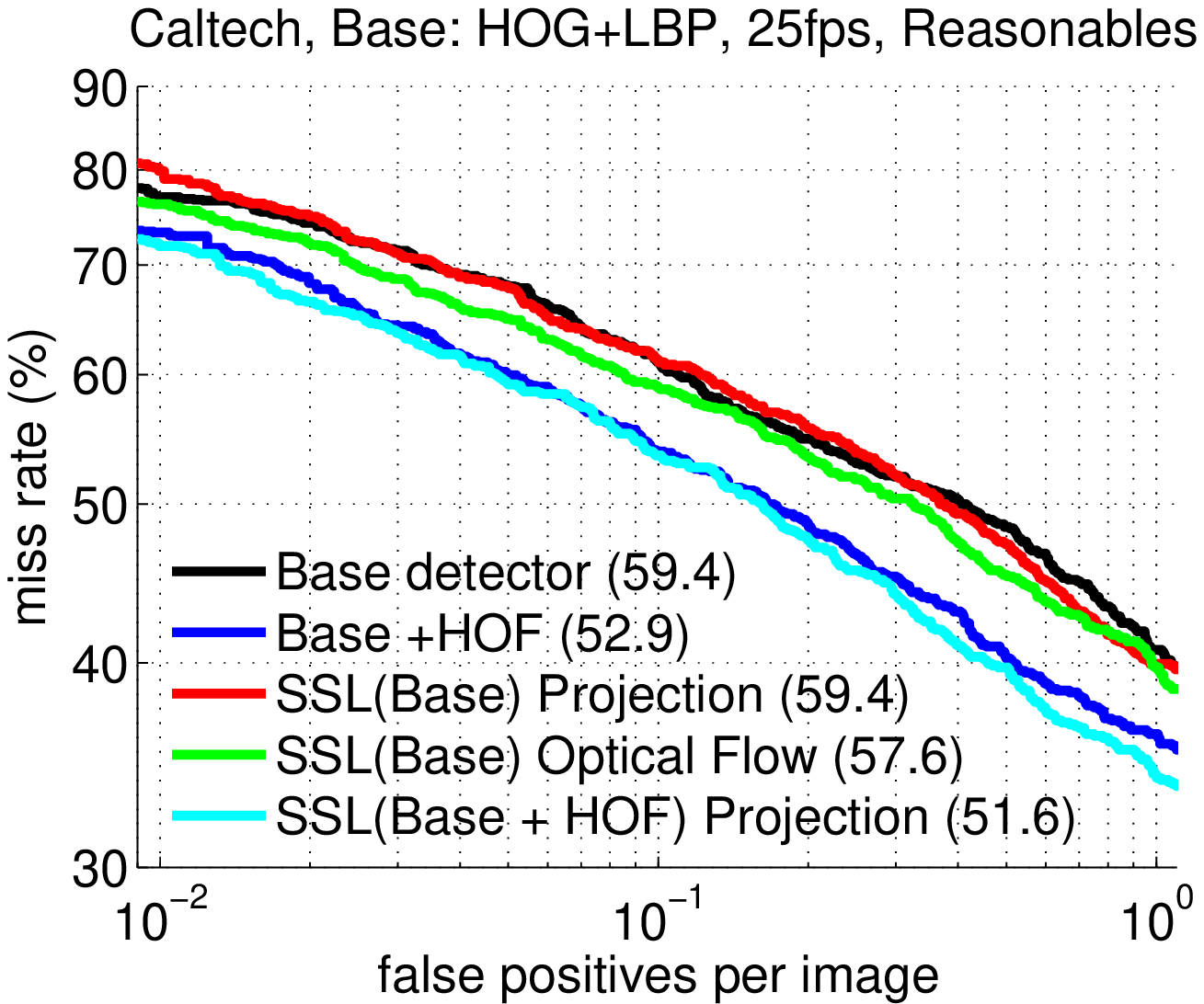}
}
%\caption{Results for different frame rates. The 10 fps and 3 fps cases are obtained by sub-sampling the video sequence, but always keeping the same training and testing pedestrians. At the bottom we also include the case of training and testing with different fps.
\caption{Results for OursDS and Caltech datasets. At the top row there are the 30fps, 10fps and 3fps cases of OursDS using the {\em near} testing subset. The last two cases are obtained by sub-sampling the video sequence, but always keeping the same training and testing pedestrians. At the bottom row there are the experiments over the {\em near}, {\em medium} and {\em reasonable} testing of Caltech dataset.
\label{fig:PlotsResults}}
\end{figure*}

\paragraph {\bf SSL.} The experiments are based on the spatiotemporal SSL (with past temporal window style) and settings $(\Delta x, \Delta y, \Delta f)=(3, 3, 5)$. In preliminary experiments we tested several values of $K$ (Fig. 1), ranging from $K=4$ to $K=1$. The obtained results were very similar, thus we decided to set $K=1$ (i.e., omitting the partition of the training sequence) since then the training is faster.

\begin{figure*}%[t!]
\centering
{\includegraphics[width=0.74\linewidth]{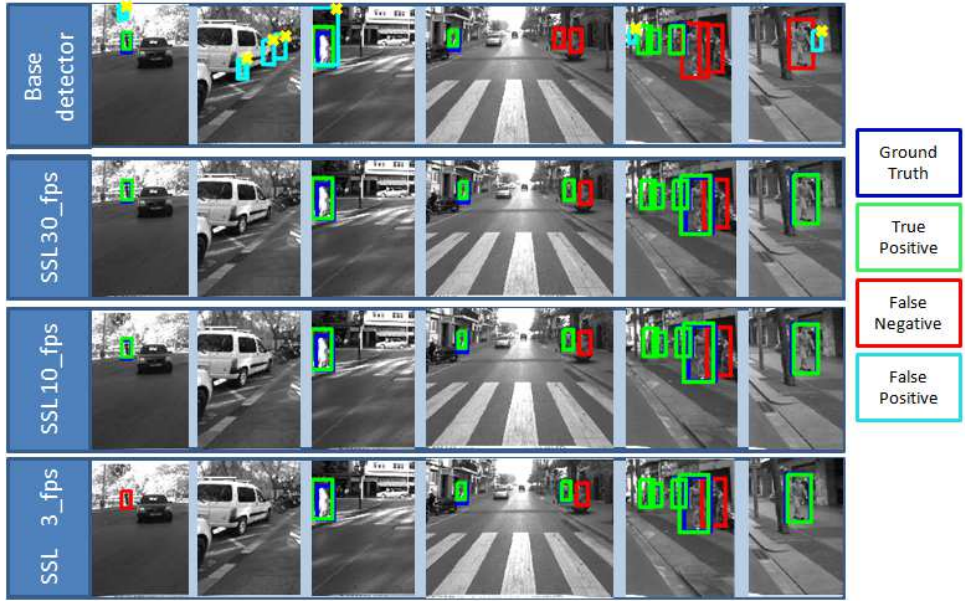}}
\caption{Qualitative results from the OursDS dataset comparing the base classifier and the SSL for 3, 10 and 30 fps. The first three columns focus on improvements regarding false positives rejection, while the rest focus on examples where SSL avoids missing pedestrians. The non-detected pedestrians with the SSL approach (last two columns) correspond to occluded pedestrians.}
\label{fig:qualitative}
\end{figure*} 

\paragraph {\bf Experiments.}  In table \ref{SSL:table:Results} we show the results for the SSL experiments. As baseline detectors we use the Base and Base+HOF. The experiments are run over the different datasets, and different frame rates for the OursDS case. We tested them for different ranges of pedestrian sizes. We observe significant accuracy improvements for all the tested datasets comparing the baseline detector and its SSL counterpart. For instance, in OurDS near with SSL(Base+HOF) we obtain an accuracy improvement of ten points approximately. Also, significant accuracy improvements are obtained for all the tested frame rates (30 fps, 10 fps, 3 fps) of OurDS dataset. Besides, we observe an improvement due to the optical flow in the volume generation at high frame rates. However, no significant difference is observed at low frame rates. The SSL accuracy improvement is more clear for the near pedestrians. In \fig{fig:PlotsResults} we plot the accuracy curves obtained for some representative experiments.

\paragraph {\bf Discussion.} SSL approach outperforms its baseline in almost all the tested configurations. However, the improvement is more clear for near pedestrians at high frame rates. If we generate the {\em past neighborhood} over the far away pedestrians, we should expect a {\em past neighborhood} with pedestrians smaller than the minimum pedestrian size that the base detector can detect. That is why the SSL improvement is not so clear for the medium subset. However, in near pedestrians {\em past neighborhood} is more probable to find a history of confident responses. This is a very relevant improvement since for close pedestrians the detection system has less time to take decisions like braking or doing any other manoeuvre. Regarding the neighborhood generation approaches, the optical flow slightly improves the projection one as it captures the movement of the pedestrians in the temporal neighborhood.

%------------------------------------------------------------------------
\section{Conclusion}
\label{sec:conclusion}

In this paper we have presented a new method for improving pedestrian detection based on spatiotemporal SSL. We have shown how even simple projection windows can boost the detection accuracy in different datasets acquired on-board. We have shown that our approach is effective for different frame rates. In this paper we have focused on HOG+LBP/Linear-SVM and HOG+LBP+HOF/Linear-SVM pedestrian base classifiers, thus, our immediate future work will focus on testing the same approach for other base classifiers of the pedestrian detection state-of-the-art. Regarding the improvement obtained using optical flow neighborhood, we want to further explore different approaches for dealing with the neighborhood generation for moving pedestrians.

%{\small
\section*{Acknowledgements}
This work is supported by the Spanish MICINN projects TRA2011-29454-C03-01 and TIN2011-29494-C03-02 and Sebastian Ramos' FPI Grant BES-2012-058280.
%}

{\small
\bibliographystyle{ieee}
\bibliography{}
}

\end{document}